\RequirePackage{fix-cm}
\documentclass[twocolumn]{svjour3}          

\usepackage{geometry}
\usepackage{multirow}
\usepackage{amssymb}
\usepackage{amsmath}
\usepackage[shortcuts]{extdash}
\usepackage{makecell}
\usepackage{xcolor}
   \definecolor{mygreen}{RGB}{121, 149, 64}
   \definecolor{myblue}{RGB}{0, 112, 192}
   \definecolor{myyellow}{RGB}{191, 144, 0}
   \definecolor{myred}{RGB}{192, 80, 70}

\smartqed  
\usepackage{graphicx}
\usepackage{booktabs}
\usepackage{url}

\begin{document}
\title{Pose-Guided Graph Convolutional Networks for Skeleton-Based Action Recognition \thanks {This work was supported by the Major Project of the Korea Institute of Civil Engineering and Building Technology (KICT) [grant number number 20210397-001].}}

\author{Han Chen \and Yifan Jiang \and Hanseok Ko}
\institute{Han Chen, \\\email{jessicachan@korea.ac.kr} 
\\ Yifan Jiang, \\\email{yfjiang@korea.ac.kr}
\\ Hanseok Ko (corresponding author), \\\email{hsko@korea.ac.kr}
\at School of Electrical Engineering, Korea University, Seoul 02841, South Korea}

\date{Received: date / Accepted: date}
\maketitle

\begin{abstract} 

Graph convolutional networks (GCNs), which can model the human body skeletons as spatial and temporal graphs, have shown remarkable potential in skeleton-based action recognition. However, in the existing GCN-based methods, graph-structured representation of the human skeleton makes it difficult to be fused with other modalities, especially in the early stages. This may limit their scalability and performance in action recognition tasks. In addition, the pose information, which naturally contains informative and discriminative clues for action recognition, is rarely explored together with skeleton data in existing methods. In this work, we propose pose-guided GCN (PG-GCN), a multi-modal framework for high-performance human action recognition. In particular, a multi-stream network is constructed to simultaneously explore the robust features from both the pose and skeleton data, while a dynamic attention module is designed for early-stage feature fusion. The core idea of this module is to utilize a trainable graph to aggregate features from the skeleton stream with that of the pose stream, which leads to a network with more robust feature representation ability. Extensive experiments show that the proposed PG-GCN can achieve state-of-the-art performance on the NTU RGB+D 60 and NTU RGB+D 120 datasets.

\end{abstract}

\section{Introduction}
\label{intro}

Human action recognition is crucial in various applications ranging from video surveillance and human-computer interaction to video understanding. In recent years, skeleton-based human action recognition has attracted significant research attention due to the development of low-cost motion sensors and their robustness when faced with complicated environments such as background clutter and changes in illumination.

Skeleton data for human recognition are made up of time sequences of 3D coordinates for human joints derived from pose estimation methods or the direct measurement by sensors, e.g., Kinect and wearable inertial measurement units \cite{hou2021self}. Early deep-learning-based action recognition methods feed the skeleton sequences into a recurrent neural network (RNN) \cite{song2017end,zheng2019relational} or employ them as a pseudo-image input for a convolutional neural network (CNN) \cite{li2018co,liang2019three,caetano2019skelemotion,caetano2019skeleton,chen2021action} to classify the action labels. To further explore the inherent correlations between human joints, graph convolutional networks (GCN)-based methods \cite{yan2018spatial,ding2019attention,li2021pose,thakkar2018part,peng2020learning,shi2020skeleton,song2021constructing,shi2019skeleton,cheng2020skeleton,ye2020dynamic,liu2020disentangling,plizzari2021skeleton,shi2020decoupled,song2020stronger,chen2021learning} have been proposed to model the natural topological structure of the human body and have achieved promising results in the human action recognition tasks. 

Despite the encouraging results achieved by previous work, state-of-the-art GCN-based methods are limited in the following aspects: 1) Scalability: The graphical form of the skeleton representation limits the fusion with other modalities, especially in the early or low-level stages, thus making it difficult to learn the features from one data stream under the supervision of the other data stream, which restricts the recognition performance. 2) Flexibility: Existing GCN-based methods mainly employ manually defined topologies to model the natural connections between human joints. However, this ignores the relationship between unnaturally connected joints such as the hands and legs, which limits the representation ability of GCNs. 3) Ignorance of pose information: Human pose data carry rich information on the spatial and temporal dynamics of human joints and have proven to be effective in several recent approaches \cite{luvizon20182d,liu2019joint,yan2019pa3d,duan2021revisiting} to action recognition, but few studies have considered utilizing pose data to enhance skeleton-based model performance, especially with the GCN-based models.

To resolve these issues, a novel pose-guided graph convolutional network (PG-GCN) is proposed in this work. To enhance the scalability of the network, instead of solely employing skeleton data as input for the GCN, we deploy a multi-stream architecture suitable for multi-modal inputs (i.e., pose and skeleton data). Furthermore, a dynamic attention module is also proposed for feature fusion in the early stages across different streams. This is achieved by employing a shared graph that bridges and refines the learned features from the skeleton data with those from the pose data. This module is trained and updated jointly with other base graph convolutional parameters within the model, thus enhancing its flexibility in constructing the graph for the skeleton. Through the multi-stream architecture and dynamic attention module, the features from the skeleton stream are aggregated with the pose information, and the robust pose-guided graph features are then used for classification, which enhances the generalization and representation ability of our proposed model.

To verify the superiority of our PG-GCN, extensive experiments are conducted on two challenging datasets: NTU RGB+D 60 and NTU RGB+D 120. The experimental results show that our model outperforms most state-of-the-art approaches. Ablation analysis of the proposed method confirms the effectiveness of the dynamic pose-guided module. The main contributions of this work are summarized below:

\begin{itemize}
\item[$\bullet$]We propose the PG-GCN, a multi-modal framework for human action recognition that can effectively fuse pose information with skeleton data and be trained end-to-end.
\end{itemize}

\begin{itemize}
\item[$\bullet$]We propose a dynamic pose-guided attention module (PG-AM) that employs a trainable shared graph to extract and fuse features across multi-stream inputs, providing more powerful graph modeling capabilities and generalization.
\end{itemize}

\begin{itemize}\item[$\bullet$]We conduct extensive experiments to show that the proposed PG-GCN outperforms state-of-the-art methods on two skeleton-based action recognition benchmarks, NTU RGB+D 60 and NTU RGB+D 120.
\end{itemize}

\section{Related Work}
\label{Realted work}

\noindent \textbf{Skeleton-based action recognition.} 
With the development of deep learning, data-driven methods have become widespread for human action recognition. Some early studies utilized RNNs or CNNs to learn the temporal dynamics of skeleton sequences \cite{song2017end,zheng2019relational,li2018co,liang2019three,caetano2019skelemotion,caetano2019skeleton,chen2021action}. However, these methods failed to represent the structure of the skeleton data, which are naturally embedded in graphs. Recently, GCNs have been widely adopted for skeleton-based action recognition \cite{yan2018spatial,ding2019attention,li2021pose,thakkar2018part,peng2020learning,shi2020skeleton,song2021constructing,shi2019skeleton,cheng2020skeleton,ye2020dynamic,liu2020disentangling,plizzari2021skeleton,shi2020decoupled,song2020stronger,chen2021learning} due to their ability to represent relationships between human body joints. The first attempt at a GCN for action recognition was ST-GCN \cite{yan2018spatial}, in which the spatial graph convolutions and temporal convolutions were combined for spatio-temporal modelling. Following this work, Liu et al. \cite{liu2020disentangling} proposed MS-G3D and explored the effects of a multi-adjacency GCN for action recognition. However, despite the success of GCNs in skeleton-based action recognition, most current methods employ a topology that is pre-defined according to the human body structure, and this topology is fixed in both the training and testing phases, which limits the generalization ability of the model. 

To further explore discriminative features and boost the performance of the skeleton-based action recognition models, efforts have been made to extract patterns from other modalities. However, most methods focus on the fusion with modalities such as RGB and depth, and few of them have considered the pose information, which carries rich information on the spatial and temporal dynamics of human joints. A recent work \cite{duan2021revisiting} took advantage of pose estimation results and verified their effectiveness in action recognition. This method embedded temporal pose estimation results as a 3D feature representation and then sent it to a 3D CNN to learn the spatio-temporal features. However, this model failed to explicitly exploit the relationship between the pose information and skeleton sequences.

\begin{figure*}
\centering
\includegraphics[width=17cm]{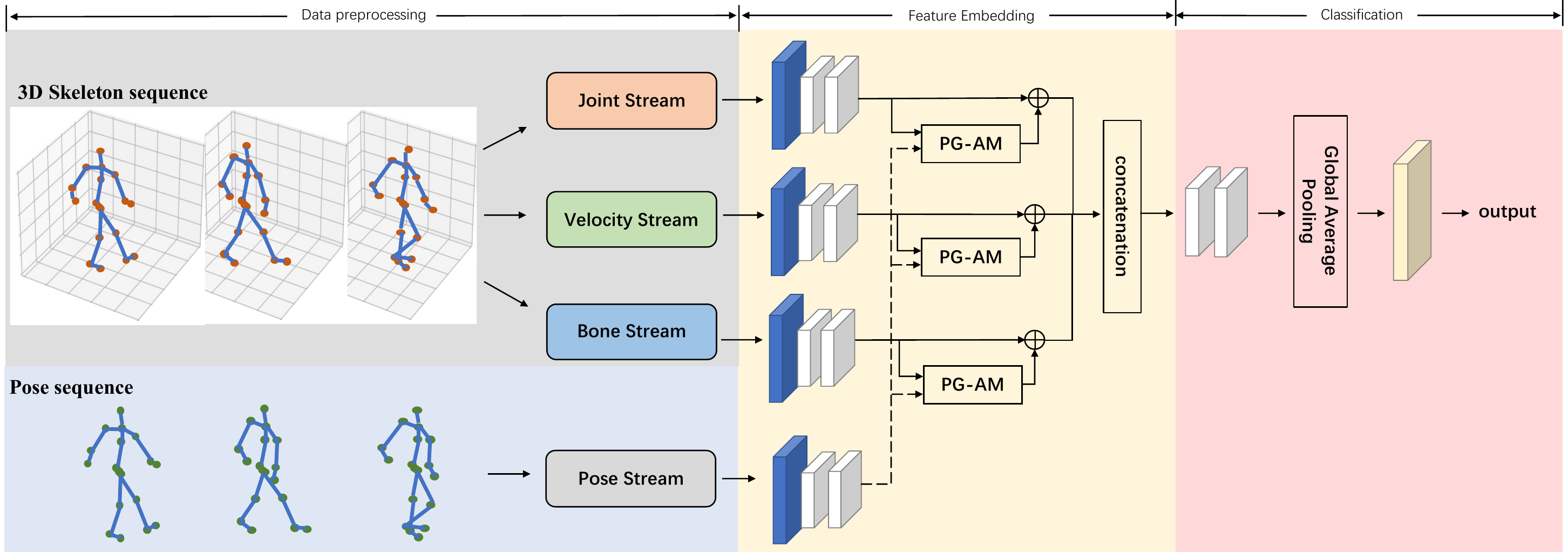}
\caption{Overview of the proposed PG-GCN model. A pair of skeleton sequence and pose sequence from the same action fragment $\{I_s, I_p\}$ are first pre-processed and then fed into the feature embedding module to obtain the feature representations $\{F_s, F_p\}$. Then, the pose-guided attention module (PG-AM) computes the attention summaries that encode the correlations between $F_s$ and $F_p$. Finally, the skeleton graph representation encoded with pose information will be handed over to the classification module to produce the action classification predictions.}
\label{fig:overview}       
\end{figure*}

\noindent \textbf{Pose for action recognition.}
Pose coordinates and skeleton data are closely related information because both are concerned with the understanding of human motion. Recently, some research has proven the effectiveness of pose information for action recognition. Yan et al. \cite{yan2018spatial} used OpenPose \cite{cao2017realtime} to extract the pose from each frame and then tested their skeleton-based action recognition. Liu et al. \cite{liu2018recognizing} also proposed the utilization of pose heatmaps estimated from RGB images input to enhance the skeleton-based action recognition. Despite the potential advantages of pose information, it has rarely been employed as top-stream input for GCNs or 3D CNNs. Recently proposed pose-based action recognition algorithms are less powerful and not sufficiently robust against the noise present during the pose estimation process. Some studies \cite{luvizon20182d,liu2019joint,yan2019pa3d} have attempted to solve both action recognition and pose estimation at the same time with a multi-task framework, confirming that pose features can be used for action recognition. To the best of our knowledge, however, pose and skeleton data have rarely been explored simultaneously in skeleton-based action recognition, especially with the GCN models. This work is thus the first to explore the use of pose data as guide information instead of only using them as the sole input for GCN.

\section{Proposed Algorithm}
\label{Proposed Algorithm}
Our PG-GCN formulates action recognition as a pose-guided graph representation learning process. The pose-guided attention module (PG-AM) learns to explicitly encode correlations between the pose and skeleton from the same sequence, enabling PG-GCN to fuse multi-stream inputs, thus further helping to discover the generalized features and producing more robust recognition results. Specifically, during training, the pose-guided procedure can be decomposed into correlation learning between the learned graph feature pairs from the same sequence (Fig. \ref{fig:overview}). During testing, the PG-GCN takes advantage of the pose-guided attention information between the pose and skeleton input. We elaborate on the pose-guided attention mechanism in Section \ref{pose-guided attention} and detail the overall PG-GCN architecture in Section \ref{full pg-gcn architecture}.

\subsection{Pose-guided Attention Module in the PG-GCN}
\label{pose-guided attention}

\begin{figure}
\centering
\includegraphics[width=8.5cm]{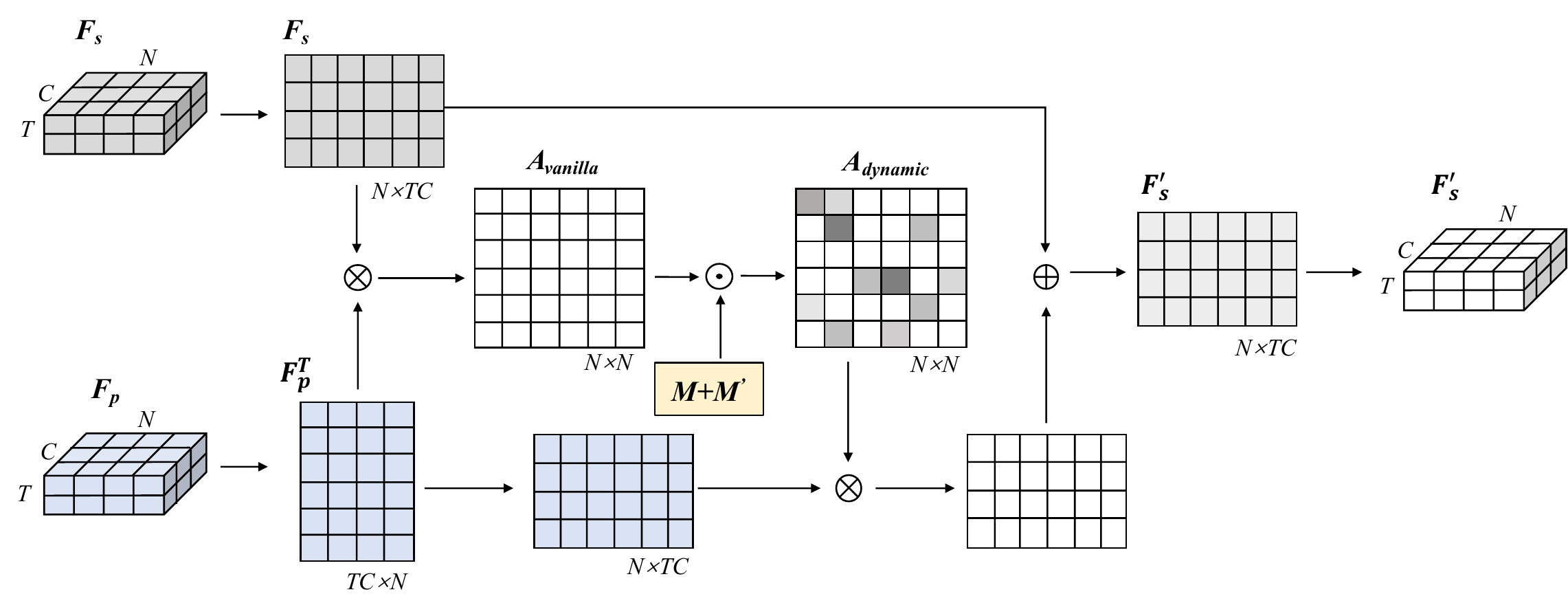}
\caption{Illustration of our PG-AM. The yellow box indicates that the parameter is learnable. $\otimes$ denotes the matrix multiplication. $\odot$ denotes element-wise multiplication. $\oplus$ denotes the element-wise summation.}
\label{fig:pg-fs}       
\end{figure}

\textbf{Vanilla pose-guided attention.} As shown in Fig. \ref{fig:pg-fs}, the two types of inputs are 2D pose sequence ${I_p}$ and 3D skeleton sequence ${I_s}$ from the same action fragment. ${F_p\in\mathbb{R}^{T\times{N}\times{C}}}$ and ${F_s\in\mathbb{R}^{T\times{N}\times{C}}}$ denote the corresponding feature representations from the feature-embedding network. ${F_p}$ and ${F_s}$ are 3D tensors with ${C}$ channels, ${T}$ frames and ${N}$ joints. The proposed pose-guided attention mines the correlations between ${I_p}$ and ${I_s}$ in the feature-embedding space. This is achieved using ${A\in[0,1]}$ to convert features from one input stream to another. For example, the features of the hand joint in the pose feature map can potentially guide the feature learning for the arm joint in the skeleton graph. To achieve pose-guided attention from ${I_p}$ to ${I_s}$, we first compute the affinity matrix between ${F_p}$ and ${F_s}$,
\begin{equation}
\left.A_{vanilla}=softmax(F_{s}F_{p}^{T})\in\mathbb{R}^{N\times{N}},\right.
\label{eq1}
\end{equation}
\noindent where $F_{p}\in\mathbb{R}^{N\times({TC})}$ and $F_{s}\in\mathbb{R}^{N\times({TC})}$ are flattened into a matrix representation. As a result, each entry for ${Att}$ reflects the similarity between the features of each joint in ${F_p}$ and ${F_s}$. 

\noindent\textbf{Dynamic pose-guided attention.} Furthermore, we propose the trainable affinity matrix $A_{dynamic}$, which is also an ${N\times{N}}$ matrix. In contrast to the vanilla affinity matrix in Eq. \ref{eq1}, the elements of the dynamic affinity matrix are parameterized and optimized together with the other parameters in the training process. $A_{dynamic}$ is first initialized by $A_{vanilla}$, modeling a prior for the correlation between the features of the pose stream and skeleton streams. Using this adjustable affinity matrix, the model can explore the most beneficial features of the recognition task. Dynamic pose-guided attention is formulated as follows,
\begin{equation}
\left.A_{dynamic}=A_{vanilla}\odot{(M+M^{'})},\right.
\label{eq2}
\end{equation}
\noindent where $M, M^{'}\in\mathbb{R}^{N\times{N}}$ denotes the trainable parameters and all of their elements are initialized with $1, 0$ respectively. In addition, $\odot$ denotes element-wise multiplication. Thus, during the training process, each element in $A_{dynamic}$ is adaptively tuned to capture a flexible correlation between the pose feature and skeleton features.

After obtaining the affinity matrix $A_{dynamic}$, we use it to fuse the pose features to the skeleton features. Given the pose feature $F_p$ from one action sequence, the skeleton feature is updated as,
\begin{equation}
\left. F_s^{'}=A_{dynamic}F_p+F_s\in\mathbb{R}^{T\times{N}\times{C}}, \right.
\label{eq3}
\end{equation}
Thus, the feature of each joint in the skeleton stream adaptively absorbs detailed information from ${F_p}$. The fused ${F_s^{'}}$ is fed into the mainstream to produce a final action classification result. 

\subsection{Full PG-GCN Architecture}
\label{full pg-gcn architecture}

The pipeline of our proposed PG-GCN is presented in Fig. \ref{fig:overview}. The PG-GCN is fundamentally a framework that consists of three cascaded components: an ST-GCN-based \cite{yan2018spatial} feature-embedding module, a pose-guided attention module (detailed in Section \ref{pose-guided attention}), and a classification module. Inspired by \cite{song2022constructing}, in which the joint positions, motion velocities, and bone features (i.e., lengths and angles) are considered, we employ the same data preprocessing for the skeleton data to produce the input for three skeleton sub-streams to fully exploit the skeleton information. The learned features of each of the three sub-stream are sent to the attention module and the correlation with the pose features is calculated. Finally, the three sub-streams are fused and passed through the classification module.

The feature embedding module is formed by orderly stacking a batch normalization layer for fast convergence, a block implemented by the ST-GCN layer, and two GCN blocks for informative feature extraction. After this module, the pose-guided module is employed to fuse features from the pose and skeleton streams. The pose-encoded skeleton feature maps are then sent into the classification module, which consists of two GCN blocks, a global average pooling layer, and a fully connected layer.

\section{Experiments}
\label{Experiments}
\subsection{Experimental Setup}

We conduct experiments on two large-scale datasets for action recognition: NTU RGB+D 60 \cite{shahroudy2016ntu} and NTU RGB+D 120 \cite{liu2019ntu}. Ablation analysis is also employed to evaluate the contribution of each component in our proposed PG-GCN.

\noindent \textbf{NTU RGB+D 60.} This large-scale dataset has been widely used to evaluate action recognition models. It contains 56,000 action clips for 60 action classes. The clips feature 40 volunteers ranging from 10 to 35 years old. Each clip is captured by three Kinect cameras from different angles: ${-45^{\circ}}$, ${0^{\circ}}$, and ${45^{\circ}}$ angles simultaneously. This dataset provides 3D skeleton sequences and corresponding 2D pose coordinates. A total of 25 human joints are captured. The author of this dataset recommends two benchmarks: 1) Cross-Subject (X-Sub), where half of the 40 volunteers are used for training (40,320 videos) and the rest for testing (16,560 videos); and 2) Cross-View (X-View), in which the sequences captured by cameras 2 and 3 are used for training (37,920 videos), and those captured by camera 1 are used for testing (18,960 videos). 

\noindent \textbf{NTU RGB+D 120.} This is an extension of NTU RGB+D 60 and is currently the largest indoor action recognition dataset. It contains 114,480 action clips for 120 classes. The clips feature 106 volunteers. It also provides 3D skeleton sequences and corresponding 2D pose coordinates. Similarly, two benchmarks are suggested for this dataset: 1) Cross-Subject (X-Sub120), in which half of the 106 subjects are used for training (63,026 videos) and the rest for testing (50,922 videos); and 2) Cross-Setup (X-Setup120), in which the training (54,471 videos) and testing (59,477 videos) sets are split based on the parity of the camera setup IDs.

\noindent \textbf{Implementation details.} 
All experiments are conducted on the PyTorch deep learning framework \cite{paszke2017automatic}. Stochastic gradient descent (SGD) is applied as the optimization strategy with a learning rate of 0.1. The batch size is set at 16. Cross-entropy is employed as the loss function. The weight decay is set at 0.0001. The network training is accelerated with an NVIDIA RTX 3090 and an Intel(R) Core i7-9700K CPU.

In the NTU RGB+D 60 and NTU RGB+D 120 datasets, there are at most two people in each clip. We pad the data for the second person with 0 if there are fewer than 2 people in the clip. The maximum number of frames in each clip is 200. Sequences with fewer than 200 frames are padded with 0 at the end. In the experiments for X-View, a transformation \cite{shi2019two} is conducted for view alignment.

\setlength{\tabcolsep}{3mm}{
\begin{table}[t]
\begin{center}
\begin{tabular}{l|c|cc}
\hline
Method & Year & X-Sub & X-View\\
\hline
STA-LSTM \cite{song2017end} & 2017 & 73.4 & 81.2 \\
ARRN-LSTM \cite{zheng2019relational} & 2018 & 81.8 & 89.6 \\
HCN \cite{li2018co} & 2018 & 86.5 & 91.1 \\
3SCNN \cite{liang2019three} & 2019 & 88.6 & 93.7 \\
\hline
ST-GCN \cite{yan2018spatial} & 2018 & 81.5 & 88.3 \\
AR-GCN \cite{ding2019attention} & 2019 & 85.1 & 93.2 \\
PR-GCN \cite{li2021pose} & 2020 & 85.2 & 91.7 \\
PB-GCN \cite{thakkar2018part} & 2018 & 87.5 & 93.2 \\
GCN-NAS \cite{peng2020learning} & 2019 & 89.4 & 95.7 \\
JB-AAGCN \cite{shi2020skeleton} & 2019 & 89.4 & 96.0 \\
EfficientGCN-B0 \cite{song2021constructing} & 2021 & 89.9 & 94.7 \\
DGNN \cite{shi2019skeleton} & 2019 & 89.9 & 96.1 \\
4s Shift-GCN \cite{cheng2020skeleton} & 2020 & 90.7 & \textbf{96.5} \\
Dynamic GCN \cite{ye2020dynamic} & 2020 & 91.5 & 96.0 \\
MS-G3D Net \cite{liu2020disentangling} & 2020 & 91.5 & 96.2 \\
\hline
PG-GCN (Ours) & 2022 & \textbf{91.8} & 95.8 \\
\hline
\end{tabular}
\end{center}
\caption{Comparison with state-of-the-art methods on NTU RGB+D 60 dataset with Top-1 accuracy (\%). The first section shows RNN or CNN-based methods, while the second section includes GCN-based models.}
\label{tab:ntu-sota}
\end{table}}


\subsection{Comparison with State-of-the-Art Methods}
\label{exp-sota}
We compare our proposed method with the state-of-the-art skeleton-based action recognition methods on both the NTU RGB+D 60 dataset and NTU RGB+D 120 datasets. The comparison methods include RNN-based \cite{song2017end,zheng2019relational}, CNN-based \cite{li2018co,liang2019three,caetano2019skelemotion,caetano2019skeleton,duan2021revisiting}, and GCN-based methods \cite{yan2018spatial,ding2019attention,li2021pose,thakkar2018part,peng2020learning,shi2020skeleton,song2021constructing,shi2019skeleton,cheng2020skeleton,ye2020dynamic,liu2020disentangling,plizzari2021skeleton,shi2020decoupled,song2020stronger,chen2021learning}. 

\noindent \textbf{NTU RGB+D 60 dataset.} Table \ref{tab:ntu-sota} presents a summary of the comparisons between the proposed method and other approaches. Compared with ST-GCN \cite{song2017end}, which is currently the most widely used backbone model for skeleton-based action recognition, our PG-GCN exhibits an improvement of over 10\% on X-Sub and 7\% on X-View. PR-GCN \cite{li2021pose} also utilizes pose information to enhance a skeleton-based action recognition model, with the pose data treated as prior information in refining the input skeleton information to reduce the impact of noise. The proposed method also outperforms PR-GCN for both benchmarks. In addition, AR-GCN \cite{ding2019attention}, JB-AAGCN \cite{shi2020skeleton}, and Dynamic GCN \cite{ye2020dynamic} also employ an attention mechanism, but there are obvious differences between these models and our PG-GCN, e.g., our attention is achieved through pose guidance, while these models focus on selecting key joints or introducing semantic information. JB-AAGCN attempts to learn a dynamic graph topology in a data-driven manner, leading it to perform better than our PG-GCN on X-View but significantly worse on X-Sub.

Overall, our model achieves more competitive results than the state-of-the-art models, confirming the superiority of our model. Notably, our method is the first to utilize pose data to guide and train a skeleton-based action recognition model, effectively maximizing the use of the pose information and benefiting the action recognition performance.

\noindent \textbf{NTU RGB+D 120 dataset.} Table \ref{tab:ntu120-sota} presents the experimental results for our proposed model and state-of-the-art methods. Of these methods, DSTA-Net \cite{shi2020decoupled}, PA-ResGCN-B19 \cite{song2020stronger}, and DualHead-Net \cite{chen2021learning} are enhanced by an attention mechanism, with the first two models exploring the dependencies between different joints in the skeleton sequence and the third utilizing attention to allow communication between coarse and fine-grained skeleton streams. Our proposed method outperforms DualHead-Net by 0.2\% on X-Sub120 and achieves competitive performance compared with the other models, which can be attributed to the utilization of pose information and the pose-guided fusion strategy.

\setlength{\tabcolsep}{2mm}{
\begin{table}[t]
\begin{center}
\begin{tabular}{l|c|cc}
\hline
Method & Year & X-Sub120 & X-Set120 \\
\hline
SkeleMotion \cite{caetano2019skelemotion} & 2019 & 62.9 & 63.0 \\
TSRJI \cite{caetano2019skeleton} & 2019 & 82.7 & 85.0 \\
PoseC3D \cite{duan2021revisiting} & 2021 & 86.9 & 90.3 \\
\hline
ST-TR-agcn \cite{plizzari2021skeleton} & 2020 & 65.5 & 59.7 \\
EfficientGCN-B0 \cite{song2021constructing} & 2021 & 85.9 & 84.3 \\
4s Shift-GCN \cite{cheng2020skeleton} & 2020 & 85.9 & 87.6 \\
DSTA-Net \cite{shi2020decoupled} & 2020 & 86.6 & 89.0 \\
MS-G3D Net \cite{liu2020disentangling} & 2020 & 86.9 & 88.4 \\
PA-ResGCN-B19 \cite{song2020stronger} & 2020 & 87.3 & 88.3 \\
DualHead-Net \cite{chen2021learning} & 2021 & 88.2 & \textbf{89.3} \\
\hline
PG-GCN (Ours) & 2022 & \textbf{88.4} & 88.8 \\
\hline
\end{tabular}
\end{center}
\caption{Comparison with state-of-the-art methods on the NTU RGB+D 120 dataset with Top-1 accuracy (\%). The first section shows RNN or CNN-based methods, while the second section includes GCN-based models.}
\label{tab:ntu120-sota}
\end{table}}

\setlength{\tabcolsep}{1mm}{
\begin{table*}[t]
\begin{center}
\begin{tabular}{l|cccc}
\hline
Method & X-Sub & X-View & X-Sub120 & X-Set120 \\
\hline
Pose-only & 88.1 & 90.9 & 82.7 & 84.2 \\
Skeleton-only & 91.4 & 95.6 & 88.2 & 88.4 \\
Pose${+}$Skeleton (w/o attention) & 90.1 & 94.3 & 85.5 & 85.5 \\
\hline
Pose${+}$Skeleton (Dynamic attention) & \textbf{91.8} & \textbf{95.8} & \textbf{88.4} & \textbf{88.8} \\
\hline
\end{tabular}
\end{center}
\caption{Comparison of different inputs on NTU RGB+D 60 and NTU RGB+D 120 with Top-1 accuracy (\%). This experiment evaluates the effectiveness of pose data for action recognition and emphasizes the need to utilize pose data to improve recognition performance.}
\label{tab:ablation-pose}
\end{table*}}

\setlength{\tabcolsep}{1mm}{
\begin{table*}[t]
\begin{center}
\begin{tabular}{l|cccc}
\hline
Method & X-Sub & X-View & X-Sub120 & X-Set120 \\
\hline
w/o attention & 90.1 & 94.3 & 85.5 & 85.5 \\
Vanilla attention & 91.6 & \textbf{95.9} & 88.3 & 88.6 \\
\hline
Dynamic attention & \textbf{91.8} & 95.8 & \textbf{88.4} & \textbf{88.8} \\
\hline
\end{tabular}
\end{center}
\caption{Comparison of different pose-guided attention mechanisms on NTU RGB+D 60 and NTU RGB+D 120 with Top-1 accuracy (\%). We also report the performance when excluding the attention module in our network.}
\label{tab:ablation-dynamic}
\end{table*}}

\subsection{Ablation Analysis}

In this section, we focus on exploration analysis of the PG-GCN components and verify the necessity of our dynamic attention strategy. The experiments are performed on the test set of NTU RGB+D 60 and NTU RGB+D 120. The evaluation criterion is the Top-1 accuracy. 

\noindent \textbf{Effectiveness of introducing pose for skeleton-based action recognition.} We first study the effect of the different data modalities for action recognition. In Table \ref{tab:ablation-pose}, we show the results when using the pose, skeleton, or pose+skeleton data. For the pose+skeleton data, the learned features from the pose stream are directly concatenated with the skeleton feature maps, serving as a separate stream in the feature-embedding module. The results show that using only pose data can lead to successful action recognition, but the performance is less competitive than just only skeleton data for training. When using both the pose and skeleton data, we observe a significant drop in performance compared with the skeleton-only model. This indicates that, even though the pose information can contribute to action recognition, simply employing it with the skeleton data as input does not lead to better performance. In contrast, our PG-GCN outperforms the other approaches across the four benchmarks, which confirms the importance of our proposed pose-guided fusion strategy. We attribute this to the dynamic pose-guided attention mechanism, which reduces the feature redundancy of the pose data while preserving the discriminative features, which benefits the learning of the features from the skeleton data.

\noindent \textbf{Effectiveness of the pose-guided attention mechanism.} We also study the effect of different pose-guided attention mechanisms in the PG-GCN, i.e., vanilla pose-guided attention (Eq. \ref{eq1}) and dynamic pose-guided attention (Eq. \ref{eq3}). As shown in Table \ref{tab:ablation-dynamic}, the dynamic attention achieves better performance than the vanilla attention mechanism. This confirms the importance of the learnable affinity matrix in dynamic attention. Furthermore, we observe a significant reduction in performance when excluding the attention module and simply concatenating the feature maps from the pose and skeleton streams (X-Sub: $91.8 \rightarrow 90.1$). The results clearly verify the effectiveness of our strategy, which employs the attention mechanism to incorporate pose information in the skeleton-based model and allows the model to learn more distinguishable features.

\begin{figure}[!t]
\centering
\includegraphics[width=8.5cm]{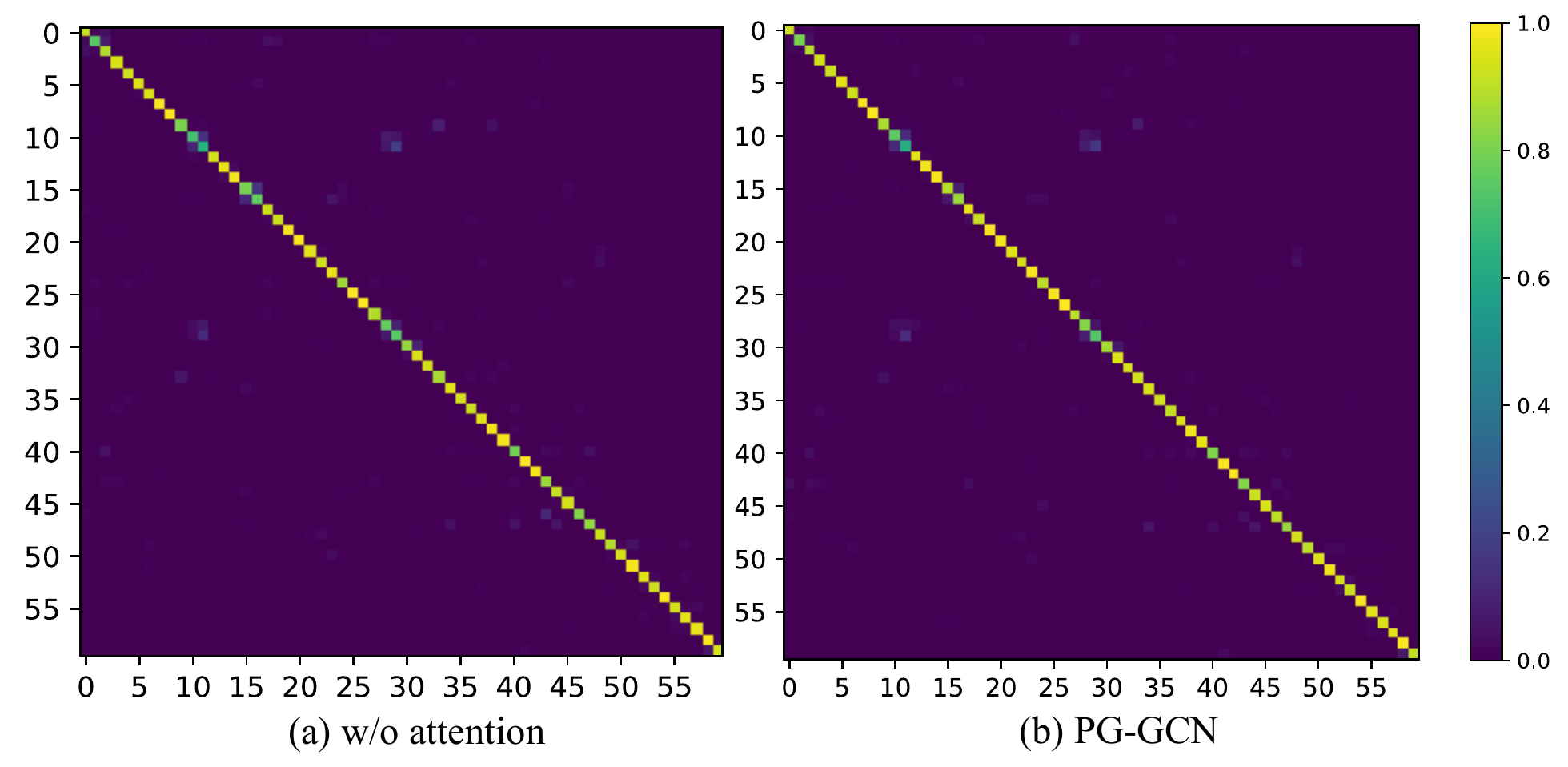}
\caption{Confusion matrix for (a) our network without the use of attention and (b) our PG-GCN with dynamic attention on the X-Sub benchmark of NTU RGB+D 60 dataset. The x-axis (true class) and y-axis (predicted class) are associated through the indices of action classes.}
\label{confusion_matrix} 
\end{figure}

\noindent \textbf{Analysis of classification confusion matrix.} To further explore the performance of our proposed method for each action class and evaluate the effectiveness of our proposed pose-guided attention mechanism, we visualize the confusion matrix on NTU RGB+D 60 (Fig. \ref{confusion_matrix}). The diagonal represents the correct classification for each action class. The non-diagonal presents the misclassification results across different action classes. Compared with the results for our network without the use of attention, the confusion matrix of our proposed method is cleaner. In other words, our proposed method achieves more accurate predictions and fewer misclassifications. This success can be attributed to our pose-guided attention mechanism, through which our network can utilize pose information to guide the robust feature learning of the skeleton. However, there are still some cases of failure in our results. For instance, the reading action (11) is often classified as playing with a phone (29), which can be attributed to the fact that these two actions include similar movements and are often confused when using sparse skeleton information. We plan to consider RGB data as complementary information to resolve these types of misclassification in future work.


\section{Conclusion}
\label{Conclusion}

In this paper, we proposed PG-GCN, a novel pose-guided multi-model framework for skeleton-based action recognition. We novelly employed the pose information as part of the input to the GCN. To fuse the features of the pose and skeleton streams, we proposed a pose-guided attention module to capture the correlations of the joints in the pose feature map and skeleton feature map as a dynamical guide for learning the graph features. The pose-guided attention module helps the network learn the most discriminating features from the skeleton sequence and improves the overall modeling capability. The proposed method achieved competitive performance on two large-scale action recognition datasets. The experimental results confirmed that our proposed method could effectively leverage pose information to improve action recognition accuracy.

\bibliographystyle{unsrt}
\bibliography{ref}
\end{document}